%% file: main.tex
\documentclass[10pt,twocolumn,letterpaper]{article}

\usepackage{cvpr}%

\input{preamble}

\definecolor{cvprblue}{rgb}{0.21,0.49,0.74}
\usepackage[pagebackref,breaklinks,colorlinks,allcolors=cvprblue]{hyperref}

\def\paperID{*****}%
\def\confName{CVPR}
\def\confYear{2026}

\title{A Real-Time Bike-Pedestrian Safety System\\with Wide-Angle Perception and Evaluation Testbed for Urban Intersections}

\author{Mehmet Kerem Turkcan\\
Columbia University\\
{\tt\small mkt2126@columbia.edu}
}

\begin{document}
\maketitle
\input{sec/0_abstract}
\input{sec/1_intro}
\input{sec/1b_related}
\input{sec/2_system}
\input{sec/3_fisheye}
\input{sec/4_decision}
\input{sec/4b_simulation}
\input{sec/5_deployment}

\input{sec/6_conclusion}
{
    \small
    \bibliographystyle{ieeenat_fullname}
    \bibliography{main}
}

\end{document}

%% file: preamble.tex
\usepackage{graphicx}

\usepackage{booktabs}
\usepackage{multirow}

\usepackage{amsmath}
\usepackage{amssymb}

\usepackage{microtype}

\usepackage{algorithm}
\usepackage{algorithmic}

%% file: sec/0_abstract.tex
\begin{abstract}
Collisions between cyclists and pedestrians at urban intersections remain a persistent source of injuries, yet few systems attempt real-time warnings to unequipped road users using commodity hardware. We present a prototype collision warning system that runs on a single edge device with a wide-angle fisheye camera, producing audible and visual alerts at 30\,fps. The system makes four contributions. First, we develop a calibration pipeline for ultra-wide fisheye lenses that overcomes corner-detection failure and optimizer divergence through perspective remapping and direct bundle adjustment. Second, we combine fisheye-aware object detection with a closed-form ground-plane projection via a precomputed lookup table. Third, we introduce a design-time conformance simulation with 24 scripted hazard scenarios, stochastic size-aware detection failures, and a latency sweep showing that a first-order kinematic predictor maintains the mean warning budget above the distracted-pedestrian reaction time across realistic camera latencies. Fourth, we formalize the decision layer as a separable, auditable testbench with explicit deployment gates, contestability mechanisms, and a residual risk register. Under conformance testing with fisheye localization error, the selected pipeline configuration achieves 93.3\% sensitivity and 92.3\% specificity, with a mean warning budget of 3.3\,s. The system design was informed by community-aided design workshops. Code and replication scripts are available at \url{https://github.com/mkturkcan/bikeped}.
\end{abstract}

%% file: sec/1_intro.tex
\section{Introduction}
\label{sec:intro}

Urban intersections concentrate the most frequent and severe conflicts between pedestrians and cyclists~\cite{nhtsa2022}. While protected lanes, signal retiming, and geometric redesign reduce exposure, these interventions require years of planning and construction. A practical infrastructure-side approach is to warn vulnerable road users (VRUs) of imminent conflicts through real-time perception systems mounted at the intersection itself.

Existing approaches fall into three categories, each with significant limitations. Vehicle-side systems such as forward collision warning and autonomous emergency braking~\cite{aeb_cyclist_sim} detect threats from the vehicle's perspective but require every vehicle to be equipped, and cannot protect pedestrians from unequipped cyclists. Connected infrastructure based on cellular vehicle-to-everything communication (C-V2X)~\cite{5gaa_vru} broadcasts warnings to connected devices, but excludes pedestrians and cyclists who do not carry compatible hardware. Post-hoc video analytics~\cite{krajewski2018highd} extract traffic patterns for planning but provide no real-time intervention.

Few systems attempt real-time, infrastructure-side collision warnings to \emph{unequipped} pedestrians and cyclists at urban intersections using only commodity hardware. This paper presents a prototype that addresses this gap and establishes a design-time conformance framework for structured evaluation prior to field deployment.

Three technical challenges define the problem. First, covering an entire crosswalk from a single mounting point requires a wide-angle fisheye lens, but the resulting distortion breaks standard calibration tools and bounding box detection. Second, the decision to alert must provide a meaningful \emph{warning budget}, that is, enough time for the pedestrian to perceive the warning and begin moving, while suppressing false alarms from benign encounters such as parallel paths. Third, the system must operate at full frame rate on edge hardware under real-world conditions including variable lighting, occlusion, and camera latency.

This paper makes four contributions:

\begin{enumerate}
    \item \textbf{Ultra-wide fisheye calibration and ground-plane projection.} Standard checkerboard calibration fails on wide-angle lenses because corner detectors assume locally straight edges and the Kannala--Brandt polynomial optimizer diverges without careful initialization. We develop a pipeline that remaps fisheye images to perspective views for corner detection, then fits the equidistant model directly via bundle adjustment. The calibrated model feeds a closed-form ground-plane projection precomputed as a pixel-level lookup table at startup (\cref{sec:fisheye,sec:calibration}).

    \item \textbf{Fisheye-aware detection and real-time edge execution.} We train a YOLO model on fisheye-augmented data, achieving a 2.5$\times$ improvement in mAP over rectilinear training, and pair it with a ground-plane tracker that maintains persistent identities across detection gaps. The full pipeline runs at 30\,fps on a single Jetson AGX Orin (\cref{sec:system}).

    \item \textbf{Hazard-oriented conformance simulation.} We evaluate the pipeline against 24 scripted hazard scenarios, including non-linear cyclist trajectories, under both deterministic and stochastic detection conditions. We sweep camera latency with three kinematic predictors and show that a first-order predictor is sufficient, while a second-order predictor degrades from noise amplification. We ground the warning budget against field-measured perception-reaction times~\cite{ped_prt_gait,cyclist_prt_asce2025,aashto2018geometric} (\cref{sec:simulation}).

    \item \textbf{A separable, auditable decision testbench.} We formalize the three-stage alert pipeline as a governance artifact with explicit deployment gates, a contestability mechanism, and a residual risk register. We compare against three structural baselines and show that the pairwise historical formulation improves specificity over naive closing while preserving comparable sensitivity, and improves sensitivity over TTC at the cost of lower specificity (\cref{sec:decision}).
\end{enumerate}

The system was deployed as a prototype on an NVIDIA Jetson AGX Orin with a single fisheye camera. Community-aided design workshops informed the alert modality and the decision to expose the pipeline logic to non-technical stakeholders. \Cref{sec:system} describes the hardware and software architecture. \Cref{sec:fisheye} derives the fisheye ground projection and calibration. \Cref{sec:decision} presents the auditable decision pipeline. \Cref{sec:simulation} presents the conformance simulation, latency analysis, and stochastic evaluation. \Cref{sec:deployment} reports the prototype demonstration and stakeholder feedback.

%% file: sec/1b_related.tex
\section{Related Work}
\label{sec:related}

Recent surveys show that roadside intelligent transportation systems have converged on a common pipeline: infrastructure sensing, calibration and fusion, trajectory reasoning, and risk assessment for warnings or control. Cre\ss{} et al. review ITS systems built around roadside infrastructure, while Zhang et al. focus specifically on roadside sensor systems for vulnerable road user (VRU) protection, emphasizing calibration, fusion, trajectory prediction, and surrogate-safety evaluation \cite{cress2024roadside,zhang2025vruprotection}.

For perception and assistance at intersections, recent work has increasingly favored sensor-rich roadside configurations. Yang et al. present VENUS, an edge-AI traffic-signal assistance system for pedestrians, cyclists, and users with disabilities that integrates roadside vision, SPaT messaging, and real-time interaction \cite{yang2022venus}. Zhang et al. propose a roadside cooperative perception system that fuses multiple cameras at an intersection, underscoring the value of coverage expansion and cross-view fusion in occluded scenes \cite{zhang2023roadsidefusion}. Likewise, Mo et al. demonstrate vehicle-to-infrastructure collaboration at obstructed intersections using roadside LiDAR and V2X communication, and Fu et al. present a digital-twin framework for pedestrian safety warning at a single urban traffic intersection \cite{mo2024v2i,fu2024digitaltwin}. Park and Kee further show that intersection-mounted LiDAR can support direct pedestrian collision-avoidance logic for right-turn conflicts \cite{park2024rightturn}.

A parallel line of work targets roadside perception models and datasets rather than end-user warning logic. Zimmer et al. introduce InfraDet3D, a roadside camera--LiDAR detector deployed at a real intersection, showing the advantage of fusing elevated infrastructure sensors for broader scene understanding \cite{zimmer2023infradet3d}. The same group released the TUMTraf Intersection Dataset, which provides synchronized roadside camera--LiDAR data and 3D annotations for complex intersection maneuvers \cite{zimmer2023tumtraf}. Very recent work continues to scale this direction: the preprint MIC-BEV proposes a multi-infrastructure camera bird's-eye-view transformer designed for heterogeneous camera layouts and degraded sensing conditions, reflecting the current shift toward large-area infrastructure-camera perception \cite{zhang2025micbev}. These works, however, generally assume multiple sensors, richer calibration, or heavier bird's-eye-view fusion pipelines than a single-camera edge deployment.

Recent studies on fixed-camera behavior prediction are also closely related to roadside warning systems. Zhou et al. formulate pedestrian crossing-intention prediction directly from surveillance video for over-the-horizon safety warning \cite{zhou2024crossing}. Abdelrahman et al. extend this direction in VRUCrossSafe, which predicts crossing intentions for multiple VRU types at intersections to support safer crossing decisions \cite{abdelrahman2025vrucrosssafe}. These papers are important because they move beyond raw detection toward proactive warning, but their emphasis is on intention prediction rather than a fully integrated roadside warning stack with explicit deployment constraints and transparent rule execution.

Finally, recent camera-centric work relevant to wide-FOV sensing has focused more on improving perception under fisheye distortion than on simplifying deployment. Kim and Park improve fisheye road-object detection via spherical projection and feature concatenation, illustrating the continued importance of distortion-aware processing for wide-angle roadside views \cite{kim2022fisheye}. At the preprint level, Traffic-Net shows that a single traffic camera can support 3D monitoring, trajectory estimation, and risk analysis through auto-calibration and tracking \cite{rezaei2021trafficnet}. Relative to this literature, the present paper occupies a narrower but practically important point in the design space: a single fisheye camera on edge hardware, metric ground-plane reasoning with minimal installation parameters, explicit cyclist--pedestrian conflict logic, and a browser-based decision testbench that makes the warning policy auditable. That combination is still comparatively underrepresented in the 2021--2026 literature, where most recent systems either prioritize richer multi-sensor cooperative perception or prediction-centric models over transparent, inspectable warning logic.

%% file: sec/2_system.tex
\section{System Architecture}
\label{sec:system}

\begin{figure}[t]
  \centering
  \includegraphics[width=0.99\linewidth]{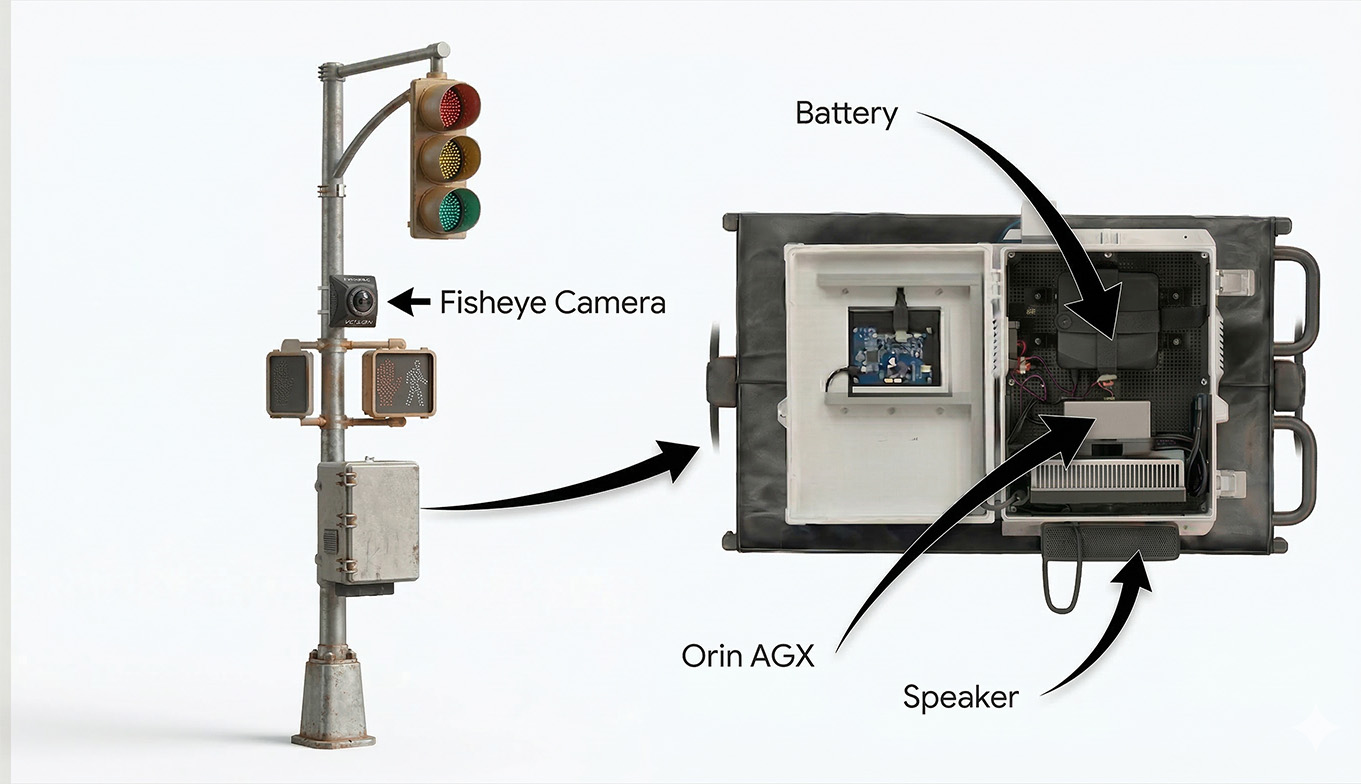}
  \caption{System overview. The deployment setup shows the fisheye camera mounted on a pole, the IP67 weatherproof enclosure, and the enclosure interior containing the Jetson AGX Orin, the programmable warning light, and the speaker.}
  \label{fig:system_overview}
\end{figure}

\begin{figure}[t]
  \centering
  \includegraphics[width=\linewidth]{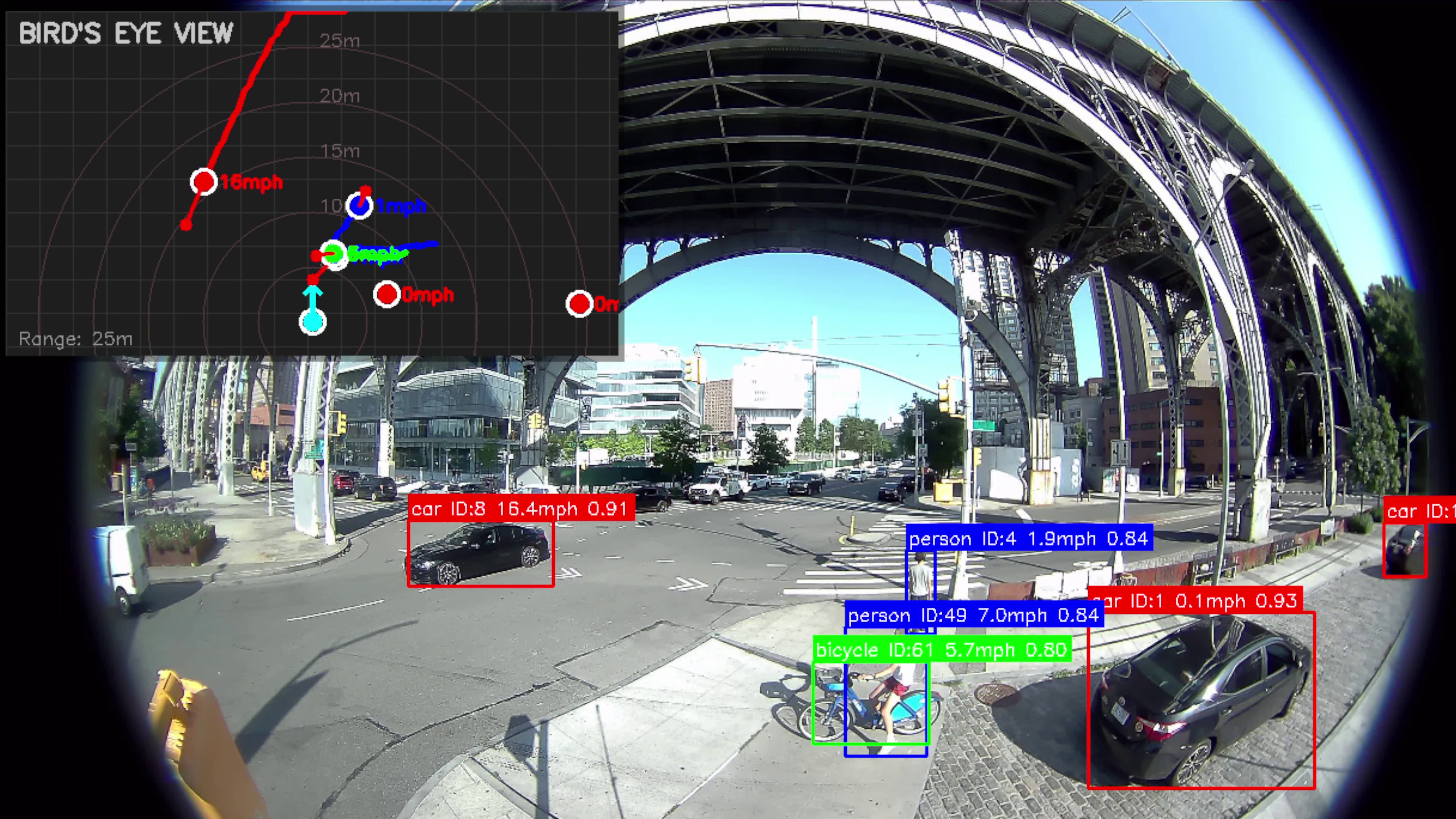}
  \caption{System output showing the fisheye camera view with detection overlays and a bird's-eye-view radar. Detected objects are projected onto a metric ground plane and displayed as a polar plot with concentric range rings at 5\,m intervals up to 25\,m.}
  \label{fig:view}
\end{figure}

The system consists of a perception module, a coordinate mapping module, a decision module, and a feedback module, all running within a single process on an NVIDIA Jetson AGX Orin. \Cref{fig:system_overview} shows the deployed hardware, and \cref{fig:view} shows the system output with bounding box overlays and a bird's-eye-view radar display.

\subsection{Hardware}

The compute unit is an NVIDIA Jetson AGX Orin with 64\,GB of unified memory. The camera captures frames at 3840$\times$2160 resolution in MJPEG format at 30\,fps through a fisheye lens. All components are housed in an IP67-rated weatherproof enclosure with a programmable warning light and speaker. The prototype was deployed at 12\,ft (3.66\,m) height with a level mounting (0$^\circ$ pitch) and a 200$^\circ$ FOV for the field demonstrations described in \cref{sec:deployment}. The deployment design-space analysis in \cref{sec:deployment_analysis} explores the parameter space for NYC traffic signal infrastructure at heights of 1.5--7.5\,m with a 220$^\circ$ FOV.

\subsection{Detection}

Object detection uses a YOLO11x model~\cite{yolo11} trained on a fisheye-augmented COCO dataset at 1280$\times$1280 input resolution. The training data was augmented with simulated fisheye distortion to match the barrel distortion characteristics of the deployed lens. The model is exported to a TensorRT engine at FP16 precision with optimization level 5. On the AGX Orin, the engine achieves a mean inference time of 21.3\,ms per frame, with 1.8\,ms for preprocessing and 0.6\,ms for postprocessing. \Cref{tab:detection} reports the detection performance on the fisheye-augmented validation set, evaluated per class and as merged groups relevant to the decision pipeline.

\begin{table}[t]
  \caption{Detection performance of the fisheye-trained YOLO11x model on the fisheye-augmented COCO validation set, reported as AP@50:95 and recall per class. Merged groups combine related classes for the decision pipeline. The original COCO-pretrained model without fisheye augmentation is shown for comparison.}
  \label{tab:detection}
  \centering
  \small
  \begin{tabular*}{\columnwidth}{@{}l@{\extracolsep{\fill}}lcc@{}}
    \toprule
    & Class & AP & Recall \\
    \midrule
    \multirow{4}{*}{\rotatebox{90}{\scriptsize Fisheye}} & Person & 0.678 & 0.755 \\
    & Bicycle + motorcycle & 0.657 & 0.747 \\
    & Car + bus + truck & 0.706 & 0.765 \\
    & All 80 classes & 0.698 & 0.771 \\
    \midrule
    \multirow{2}{*}{\rotatebox{90}{\scriptsize Orig}} & All 80 classes & 0.283 & 0.400 \\
    & Bicycle + motorcycle & 0.253 & 0.349 \\
    \bottomrule
  \end{tabular*}
\end{table}

The fisheye-trained model achieves an overall mAP@50:95 of 0.698 and recall of 0.771, compared to 0.283 and 0.400 for the original COCO-pretrained model evaluated on the same fisheye data. The confidence threshold is set to 0.1 to maximize recall of small and partially occluded VRUs, with downstream filtering handled by the tracking and decision stages. Multi-object tracking operates in the ground-plane coordinate space: each detection is projected to metric BEV coordinates via the fisheye lookup table, then matched to existing tracks by greedy nearest-neighbour within a 3\,m radius. Lost tracks persist for up to 10\,s via constant-velocity prediction. Each tracked object accumulates a position history used for speed estimation and closing-distance analysis.

\subsection{Preprocessing and Coordinate Mapping}

The raw camera frame is padded to a square and center-cropped once at startup to define the spatial extent of the ground coordinate lookup table. At runtime, the bottom center of each detected bounding box is mapped from the YOLO input space back to the preprocessed space and used to index a precomputed ground coordinate array, yielding the metric position of the detected object. The projection model that generates this array is described in \cref{sec:fisheye}.

\subsection{Tracking and Speed Estimation}

Each tracked object maintains a position history as a sequence of ground-plane coordinates indexed by frame number. Speed is estimated from the Euclidean displacement between positions separated by 4 frames, divided by the elapsed time at the camera frame rate. This windowed estimator smooths single-frame noise while remaining responsive to changes in velocity. A 4-frame window produces a usable speed estimate within 133\,ms of first detection, which is critical for high-speed approach scenarios where the cyclist traverses the alert zone in under one second.

\subsection{Communication}

The system publishes tracking data over MQTT~\cite{mqtt} at a configurable interval. Each message contains a timestamp, frame number, and a list of tracked objects with their class labels, metric positions, velocities, and position histories. This telemetry stream enables downstream consumers such as traffic management dashboards, data loggers, and the decision testbench described in \cref{sec:decision}.

%% file: sec/3_fisheye.tex
\section{Fisheye Ground Projection}
\label{sec:fisheye}

Converting pixel detections to metric ground-plane coordinates requires modeling the fisheye lens distortion and solving the ray-plane intersection with the ground. We derive a closed-form vectorized projection that precomputes the ground coordinates for every pixel in the preprocessed frame. At runtime, this lookup table converts each detection to a metric position in constant time.

\subsection{Sensor Geometry Correction}

Fisheye lenses project a circular image onto a rectangular sensor, causing the top and bottom of the scene to be cropped while the left and right margins fall outside the projection circle. This mismatch breaks the radial symmetry assumed by fisheye distortion models. The preprocessing step described in \cref{sec:system} restores approximate symmetry by padding the frame to a square and center-cropping so that the fisheye radius $R$ maps approximately uniformly in all directions. Checkerboard calibration (\cref{sec:calibration}) then recovers the true optical center $(c_x, c_y)$, which may be offset from the frame center due to the sensor--lens alignment.

\subsection{Pixel-to-Ray Conversion}

Let the preprocessed image have dimensions $W \times H$ with optical center $(c_x, c_y)$. For each pixel $(u, v)$, define the displacement from the optical center as $\Delta x = u - c_x$ and $\Delta y = v - c_y$. The radial distance and azimuth angle in the image plane are:
\begin{equation}
  r = \sqrt{\Delta x^2 + \Delta y^2}, \quad \phi = \operatorname{atan2}(\Delta y, \Delta x).
  \label{eq:polar}
\end{equation}

The radial distance $r$ is converted to the incidence angle $\theta$ through the lens distortion model. The system supports equidistant, equisolid, orthographic, and stereographic projections~\cite{kannala2006generic}. We calibrated all four models against 42 checkerboard frames captured through the fisheye lens (\cref{sec:calibration}); the equidistant model achieved 4.44\,px RMS reprojection error, within 2\% of the best-fitting equisolid model (4.34\,px), confirming the equidistant assumption. In the deployed equidistant model, the focal length $f = D \cdot 180 / (\Omega \pi)$ maps $\theta = r / f$, where $D = 2R$ is determined by the fisheye radius $R$ in pixels.

Each pixel's incidence angle $\theta$ and azimuth $\phi$ define a unit ray direction in the camera coordinate frame:
\begin{equation}
  \mathbf{d} = \begin{pmatrix} \cos\theta \\ \sin\theta\,\cos\phi \\ -\sin\theta\,\sin\phi \end{pmatrix}.
  \label{eq:ray}
\end{equation}

Here the $x$-axis points forward along the camera optical axis, the $y$-axis points to the right, and the $z$-axis points upward.

\subsection{Pitch Rotation}

The camera is mounted with a pitch angle $\alpha$ relative to the horizontal. A rotation about the $y$-axis aligns the ray directions with the world frame:
\begin{equation}
  \mathbf{d}' = \begin{pmatrix}
    \cos\alpha & 0 & -\sin\alpha \\
    0 & 1 & 0 \\
    \sin\alpha & 0 & \cos\alpha
  \end{pmatrix} \mathbf{d}.
  \label{eq:pitch}
\end{equation}

\subsection{Ray-Ground Intersection}

The camera is positioned at $\mathbf{p}_0 = (0, 0, h)^\top$ where $h$ is the mounting height above the ground plane $z = 0$. A point along the pitched ray is $\mathbf{p}(t) = \mathbf{p}_0 + t\,\mathbf{d}'$. Setting $p_z(t) = 0$ and solving for $t$:
\begin{equation}
  t = \frac{-h}{d'_z}, \quad d'_z < 0.
  \label{eq:t_param}
\end{equation}

The constraint $d'_z < 0$ ensures that only rays directed toward the ground produce valid intersections. Rays directed above the horizon yield no ground point and are masked as invalid.

The ground-plane coordinates are then:
\begin{equation}
  \mathbf{g} = \mathbf{p}_0 + t \cdot \mathbf{d}'.
  \label{eq:ground}
\end{equation}

\subsection{Precomputation}

The ground coordinate array $\mathbf{G} \in \mathbb{R}^{H \times W \times 3}$ and the validity mask $\mathbf{M} \in \{0,1\}^{H \times W}$ are computed once at system startup using vectorized NumPy operations over the full pixel grid. At runtime, converting a detection at pixel $(u, v)$ to a metric ground position requires a single array index operation: $\mathbf{g}_{u,v} = \mathbf{G}[v, u, :]$, with $\mathbf{M}[v, u] = 1$ confirming validity.

\subsection{Intrinsic Calibration}
\label{sec:calibration}

Checkerboard calibration at ultra-wide fields of view faces two algorithmic obstacles. First, standard corner detectors assume locally straight inter-corner edges, an assumption violated by the severe barrel distortion of a 200\textdegree{} lens where straight world lines curve by tens of pixels. Second, the Kannala--Brandt polynomial model~\cite{kannala2006generic} requires initial estimates of the higher-order distortion coefficients; at extreme fields of view, coarse initialization places the optimizer in a basin where it diverges rather than converging to the true projection. We address both problems with a two-stage pipeline: perspective remapping for corner detection, followed by direct bundle adjustment of the equidistant model.

In the first stage, each fisheye frame is remapped to a perspective (rectilinear) view using the approximate equidistant model, restoring locally straight edges. Corners are detected in the rectified image, mapped back to fisheye pixel coordinates, and refined to subpixel accuracy on the original frame.

The resulting 2D--3D correspondences are used in a Levenberg--Marquardt bundle adjustment that jointly optimizes the intrinsic parameters $(c_x, c_y, f)$ and per-frame extrinsics $(R_i, \mathbf{t}_i)$, minimizing reprojection error through the equidistant model directly. A second pass applies a soft-$\ell_1$ robust loss to downweight outlier corners at the fisheye periphery where the projection gradient is steepest.

The calibration reveals that the optical center is offset from the geometric frame center by $(+3.2, +55.0)$\,px, consistent with the spherical lens being mounted on a 16:9 sensor where the projection circle does not align with the sensor center. Ignoring this offset introduces systematic position errors that grow toward the periphery. Comparing all four supported projection models on the same 42-frame calibration set, the equidistant and equisolid models achieve comparable reprojection error (4.4 and 4.3\,px RMS respectively), while the stereographic and orthographic models produce substantially higher error, confirming that the equidistant model is appropriate for this lens class.

\subsection{Installation Requirements}

Beyond the calibrated intrinsics, the projection model requires three parameters from the installer: the mounting height $h$, the pitch angle $\alpha$, and the fisheye radius $R$ in pixels. The model assumes locally flat ground within the detection range. We developed calibration software that overlays the projected ground grid on the camera feed, allowing the installer to adjust the extrinsic parameters until the grid aligns with known references such as lane markings.

\subsection{Bounding Box Localization Error}
\label{sec:bbox_error}

The detector returns axis-aligned bounding boxes in the fisheye image. Because the fisheye projection warps three-dimensional objects, the bottom center of the detected bounding box does not coincide with the pixel corresponding to the true ground-contact point. This discrepancy introduces a systematic localization error that varies with object class, distance, and bearing angle.

To quantify this error, we project the eight corners of a three-dimensional bounding box through the fisheye model, compute the tightest axis-aligned rectangle enclosing the visible projected corners, and inverse-project the bottom center of this rectangle back to the ground plane. The localization error is the Euclidean distance between this estimate and the true ground position.

\Cref{tab:bbox_err} reports the error for three object classes at selected distances along the camera forward axis. Pedestrians exhibit the smallest error because their narrow cross-section produces minimal lateral shift under fisheye warping. Vehicles exhibit the largest error because their wide footprint causes the bounding box center to shift substantially from the true contact point. The error remains below 0.25\,m for pedestrians, below 0.29\,m for cyclists, and below 0.83\,m for cars at all tested distances within 25\,m.

\begin{table}[t]
  \caption{Bounding box localization error at selected distances along the camera forward axis, using the calibrated camera parameters ($f = 1013.3$\,px, $\Omega = 197.9$\textdegree).}
  \label{tab:bbox_err}
  \centering
  \small
  \begin{tabular*}{\columnwidth}{@{}l@{\extracolsep{\fill}}rrr@{}}
    \toprule
    Distance & Pedestrian & Cyclist & Car \\
    \midrule
    3\,m  & 0.248\,m & 0.223\,m & 0.387\,m \\
    5\,m  & 0.249\,m & 0.249\,m & 0.556\,m \\
    10\,m & 0.249\,m & 0.273\,m & 0.723\,m \\
    15\,m & 0.249\,m & 0.282\,m & 0.783\,m \\
    25\,m & 0.250\,m & 0.289\,m & 0.830\,m \\
    \bottomrule
  \end{tabular*}
\end{table}

At 80\textdegree{} bearing at 10\,m, pedestrian error is 0.35\,m and cyclist error is 0.89\,m, both within the proximity thresholds used by the decision pipeline. Vehicle error of 2.5\,m is larger but does not affect the pedestrian-cyclist use case.

%% file: sec/4_decision.tex
\section{Auditable Decision Pipeline}
\label{sec:decision}

The mapping from perception outputs to physical alerts constitutes the decision layer. We separate this layer from the perception stack and formalize it as a governance artifact structured around four questions:

\begin{enumerate}
\item \emph{What rule fires the alert?} The three-stage logic is fully specified in \cref{alg:decision} with five numeric parameters.

\item \emph{Where did the parameters come from?} The optimizer's cost function, bounds, and convergence trace are version-controlled alongside the scenario suite and calibration data, establishing parameter traceability: scenario suite $\to$ optimizer $\to$ selected configuration $\to$ \texttt{config.yaml} $\to$ deployed prototype.

\item \emph{Can the decision be contested?} A stakeholder may challenge a rule by proposing a new scenario, a tighter parameter bound, or a different ground-truth threshold and rerunning the conformance suite. No access to the perception stack is required. The browser testbench supports this interactively; the Python testbench supports it programmatically.

\item \emph{What gates deployment?} A configuration is promotable to \texttt{config.yaml} only if it achieves $\geq$90\% actionable-frame sensitivity across all danger scenarios with actionable frames, $\geq$90\% specificity, and a mean warning budget exceeding the distracted-pedestrian PRT (1.87\,s). These thresholds are provisional design targets, not validated acceptance criteria, and should be revisited after field evaluation.
\end{enumerate}

\subsection{Three-Stage Pipeline}

The pipeline evaluates three conditions in sequence, producing one of four output states (\cref{tab:states}):

\paragraph{Stage 1: Pedestrian presence.} If no tracked pedestrian is present, the output is IDLE.

\paragraph{Stage 2: Cyclist memory.} A temporal buffer of length $N$ frames records whether a cyclist was recently detected, accounting for brief occlusion gaps.

\paragraph{Stage 3: Pairwise closing check.} For each pedestrian-cyclist pair with ground-plane distance $d \in [d_{\min}, d_{\max}]$, the system compares the pairwise distance at frame $t$ to the distance at frame $t - k$, using the historical positions of \emph{both} agents:
\begin{equation}
  \|\mathbf{p}^{\text{c}}_t - \mathbf{p}^{\text{p}}_t\| < \|\mathbf{p}^{\text{c}}_{t-k} - \mathbf{p}^{\text{p}}_{t-k}\| \;\;\text{and}\;\; \|\mathbf{p}^{\text{c}}_t - \mathbf{p}^{\text{c}}_{t-k}\| > \Delta_{\min},
  \label{eq:proximity}
\end{equation}
where $\Delta_{\min}$ is a minimum cyclist displacement threshold over $k$ frames, filtering stationary or near-stationary cyclists. This formulation is critical. A naive implementation that compares the cyclist's past position against the pedestrian's \emph{current} position creates false convergence for co-directional paths: if both agents move in the same direction at different speeds, the distance from the cyclist's old position to the pedestrian's new position decreases even when the actual pairwise gap is constant. Using both agents' historical positions eliminates this artifact. \Cref{alg:decision} summarizes the procedure.

\begin{algorithm}[t]
\caption{Three-stage decision pipeline. $N$: cyclist memory length (frames). $k$: lookback depth (frames). $d_{\min}, d_{\max}$: proximity bounds (m). $\Delta_{\min}$: minimum cyclist displacement over $k$ frames (m).}
\label{alg:decision}
\begin{algorithmic}[1]
\REQUIRE Tracked objects $\mathcal{O}$; cyclist memory buffer $\mathcal{B}$; position history $\mathcal{H}$
\ENSURE State $s \in \{\text{IDLE}, \text{SAFE}, \text{WARNING}, \text{ALERT}\}$
\STATE $\mathcal{P} \leftarrow \{o \in \mathcal{O} \mid \text{class}(o) = \texttt{person}\}$
\STATE $\mathcal{C} \leftarrow \{o \in \mathcal{O} \mid \text{class}(o) \in \{\texttt{bike}, \texttt{motorcycle}\}\}$
\IF{$\mathcal{P} = \emptyset$}
  \RETURN IDLE
\ENDIF
\IF{$\mathcal{B}$ contains no detection within last $N$ frames}
  \RETURN SAFE
\ENDIF
\FOR{each pair $(c, p)$ with $c \in \mathcal{C},\; p \in \mathcal{P}$}
  \IF{$|\mathcal{H}[c]| \geq k{+}1$ \AND $|\mathcal{H}[p]| \geq k{+}1$}
    \STATE $d_t \leftarrow \|\mathbf{p}_c^{(t)} - \mathbf{p}_p^{(t)}\|$
    \IF{$d_{\min} \leq d_t \leq d_{\max}$}
      \STATE $d_{t-k} \leftarrow \|\mathbf{p}_c^{(t-k)} - \mathbf{p}_p^{(t-k)}\|$
      \STATE $\Delta_c \leftarrow \|\mathbf{p}_c^{(t)} - \mathbf{p}_c^{(t-k)}\|$
      \IF{$d_t < d_{t-k}$ \AND $\Delta_c > \Delta_{\min}$}
        \RETURN ALERT
      \ENDIF
    \ENDIF
  \ENDIF
\ENDFOR
\RETURN WARNING
\end{algorithmic}
\end{algorithm}

\begin{table}[t]
  \caption{Decision pipeline output states and feedback signals.}
  \label{tab:states}
  \centering
  \small
  \begin{tabular*}{\columnwidth}{@{}ll@{\extracolsep{\fill}}ll@{}}
    \toprule
    State & Light & Sound & Condition \\
    \midrule
    IDLE & Blue & None & No pedestrian detected \\
    SAFE & Green & Chime & Pedestrian, no cyclist in memory \\
    WARNING & Orange & None & Cyclist in memory, not approaching \\
    ALERT & Red blink & Bell & Cyclist approaching a pedestrian \\
    \bottomrule
  \end{tabular*}
\end{table}

\subsection{Browser-Based Testbench}

\begin{figure*}[t]
  \centering
  \includegraphics[width=0.85\linewidth]{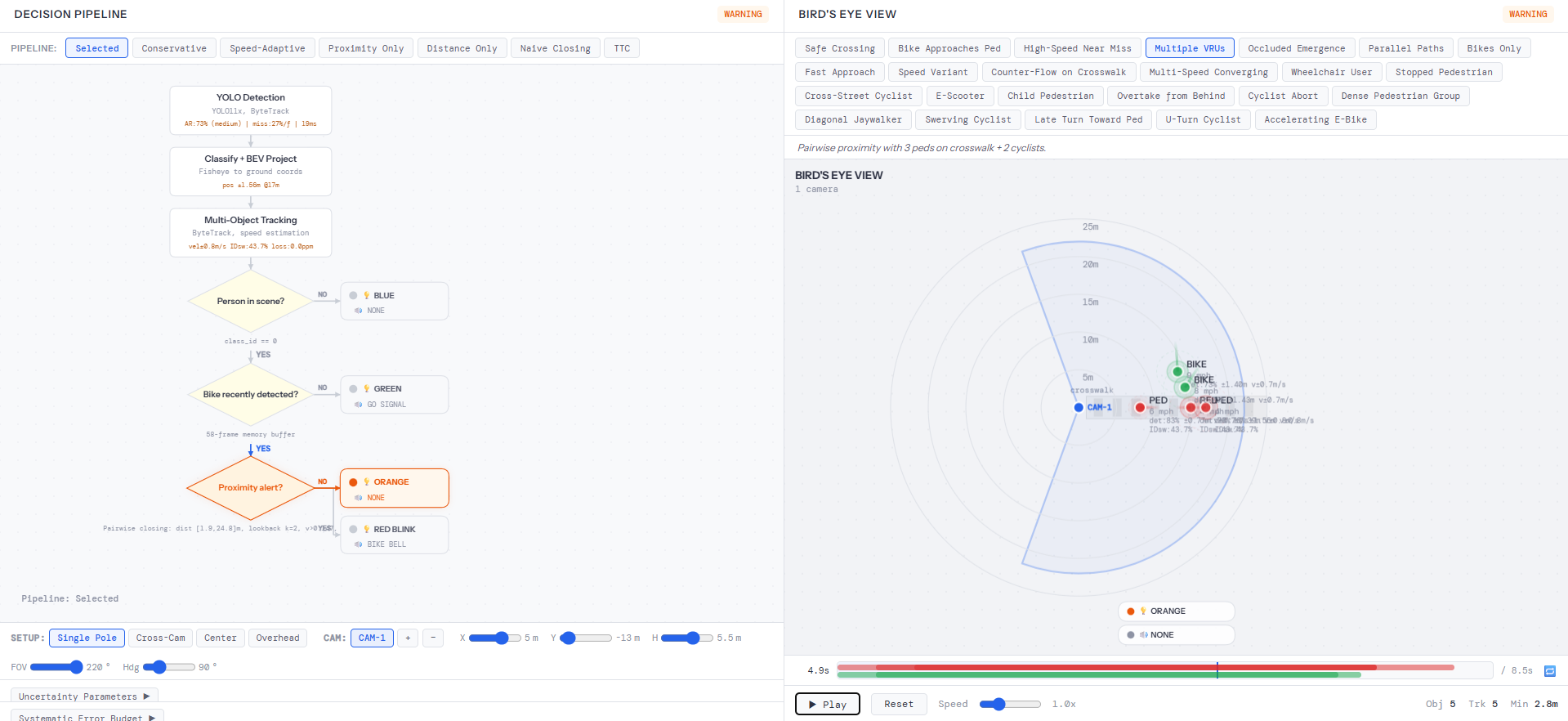}
  \caption{Browser-based decision testbench. Left: the pipeline as a flow diagram with active state. Right: bird's-eye view with agent positions, camera coverage, and range rings. Uncertainty sliders propagate detection noise in real time.}
  \label{fig:testbench}
\end{figure*}

The decision layer is packaged as a separable module with two testbench implementations: a browser-based interface for stakeholder review (\cref{fig:testbench}) and a Python testbench for parameter optimization and Monte Carlo analysis. The browser interface renders the pipeline as a flow diagram with the active state highlighted alongside a bird's-eye view of the scenario. Uncertainty sliders propagate detection noise through the pipeline in real time, allowing non-technical reviewers to explore how missed detections affect alert behavior. The Python testbench logs per-frame observations to JSON: scenario ID, configuration hash, calibration version, random seed, and for each frame the true and observed agent positions, localization error, pipeline state, ground-truth danger label, and actionability flag. This audit trail supports post-hoc review of any individual alert or missed detection.

\subsection{Decision Rule Baselines}

We compare the pairwise historical closing check against three structurally different decision rules (\cref{tab:baselines}), all using the same detection and tracking input.

\emph{Sensitivity} is the fraction of ground-truth danger frames in which the pipeline outputs ALERT. \emph{Specificity} is the fraction of safe frames without a false ALERT. \emph{SevFN} weights missed danger frames by the cyclist's kinetic energy, so that missing a high-speed threat costs more than missing a slow one. \emph{Fatigue} is the fraction of total simulation time spent in the ALERT state, measuring the risk of alert habituation.

\begin{table}[t]
  \caption{Decision rule comparison across 24 conformance scenarios with fisheye localization error. All rules use the same detection input and optimized parameters where applicable.}
  \label{tab:baselines}
  \centering
  \small
  \resizebox{\columnwidth}{!}{%
  \begin{tabular}{@{}lcccc@{}}
    \toprule
    Decision rule & Sens & Spec & SevFN & Fatigue \\
    \midrule
    Distance-only ($d < 10$\,m)        & 57.8\% & 70.2\% & 58.5\% & 41.1\% \\
    Naive closing (past vs.\ current)  & 93.4\% & 90.4\% &  7.5\% & 33.8\% \\
    TTC $< 3$\,s                       & 84.3\% & 96.4\% & 14.7\% & 27.0\% \\
    Pairwise historical (ours)         & 93.3\% & 92.3\% &  7.6\% & 32.3\% \\
    \bottomrule
  \end{tabular}}
\end{table}

The distance-only rule fires on every close pass regardless of motion, producing the worst specificity and SevFN. The naive closing check achieves sensitivity comparable to the pairwise rule but lower specificity because comparing the cyclist's past position to the pedestrian's \emph{current} position creates spurious convergence on co-directional paths. The TTC threshold achieves the best specificity but lower sensitivity, because the per-frame velocity estimate is noisy under fisheye localization error, producing unreliable TTC values. Pairwise historical improves specificity over naive closing (92.3\% vs.\ 90.4\%) while preserving comparable sensitivity (93.3\% vs.\ 93.4\%), and improves sensitivity over TTC (93.3\% vs.\ 84.3\%) at the cost of lower specificity (92.3\% vs.\ 96.4\%).

%% file: sec/4b_simulation.tex
\section{Conformance Simulation and Evaluation}
\label{sec:simulation}

\subsection{Conformance Scenario Set}
\label{sec:scenarios}

All simulation results use the deployed camera configuration: 12\,ft (3.66\,m) mounting height, 0$^\circ$ pitch (level), 197.9$^\circ$ FOV equidistant lens, with calibrated optical center ($c_x\!=\!1752.7$, $c_y\!=\!1804.5$\,px in the 3500$\times$3500 crop). The testbench loads these parameters from the same \texttt{config.yaml} and \texttt{camera\_calibration.json} files used by the deployed system and uses the identical equidistant inverse-projection to convert pixel coordinates to metric ground-plane positions. Agent positions in each scenario are specified in world coordinates; the testbench transforms them to camera-relative coordinates before entering the decision pipeline, matching the camera-centred BEV output of the deployed fisheye projection.

The results in this section constitute \emph{design-time conformance evidence}: they verify that the pipeline logic handles the enumerated encounter types under modeled perception conditions. The pipeline parameters reported in \cref{tab:pipeline_results} as ``Selected'' are the output of the differential-evolution optimizer described in \cref{sec:optimization}, run against the same scenario suite and ground-truth model used for evaluation. The optimizer finds the best parameters \emph{for} these scenarios, and the sensitivity analysis in \cref{tab:gt_sensitivity} bounds the sensitivity of the results to the ground-truth assumptions. Operational validation under real traffic, lighting, and weather conditions remains future work.

The 24 test scenarios are \emph{conformance cases}: they enumerate specific encounter types that the system must handle correctly and are not a statistical sample of real traffic. The set is designed for hazard coverage and includes: safe crossings with no cyclist present (3 scenarios), standard head-on and overtaking approaches (5), high-speed and accelerating encounters (3), accessibility cases including a wheelchair user and a child pedestrian (2), multi-agent scenes with dense pedestrian groups and multiple cyclists (3), edge cases such as cyclist abort and counter-flow on the crosswalk (3), and non-linear trajectories including swerving, late turns, U-turns, and e-bike acceleration (4). Of the 24 scenarios, 21 contain at least one ground-truth danger interval and 3 are entirely safe. Under two-tier labeling, 19 of the 21 have actionable danger frames. The remaining 2 (Occluded Emergence and Fast Approach) have all danger frames in the imminent tier where TTC $< 1.87$\,s: the cyclist is already too close for a warning to reach the pedestrian in time. These scenarios are retained in the suite because they stress-test detection at close range and high speed, and future improvements to the pipeline (e.g.\ earlier detection via higher-resolution models or multi-camera fusion) may shift their danger frames into the actionable window. The complete scenario definitions, including agent paths and timing, are provided in the supplementary codebase for reproducibility.

\subsection{Kinematic Ground Truth}
\label{sec:kinematic}

Evaluating the decision pipeline requires ground-truth labels for each frame. We use a kinematic safety model that is deliberately \emph{more conservative} than the pipeline itself: a frame is labeled dangerous if the cyclist is closing, the closest point of approach (CPA) is within 5\,m, and either the stopping distance $d_{\text{stop}} = v \cdot t_{\text{react}} + v^2 / 2a$ (with 85th-percentile field-measured values $t_{\text{react}} = 0.84$\,s and $a = 1.96$\,m/s$^2$ for conventional bicycles~\cite{cyclist_prt_asce2025}) exceeds 80\% of the remaining gap, or the time to collision (TTC) is below 3.0\,s. E-bicycle scenarios use a higher deceleration ($a = 6.0$\,m/s$^2$) reflecting disc-brake capability at 25\,km/h on dry pavement. Severity weights $s = \min(v^2/v_{\max}^2, 1)$ with $v_{\max} = 12$\,m/s assign higher cost to missed high-speed threats.

Danger frames are further classified into two tiers based on whether the system's warning can still change the outcome:
\begin{itemize}
  \item \textbf{Actionable:} TTC $\geq$ distracted-pedestrian PRT (1.87\,s). The pedestrian has time to perceive the alert and begin moving. The system is evaluated on these frames.
  \item \textbf{Imminent:} TTC $<$ 1.87\,s. The pedestrian cannot react in time. The alert is still correct (not a false positive), but the system is not penalized for missing these frames.
\end{itemize}

\noindent Sensitivity and SevFN are computed over actionable frames only. Specificity treats all danger frames (actionable and imminent) as expected-alert, so that alerting during imminent danger is not counted as a false positive. The clearance time in the TTC threshold uses the actual pedestrian speed from each scenario trajectory, not a fixed default.

The cyclist could also react to the audible alert by swerving laterally out of the conflict zone. Under constant lateral acceleration $a_{\text{lat}} = \mu g$ with conservative friction $\mu = 0.4$, the time to clear a lateral distance $w$ is $t_{\text{maneuver}} = \sqrt{2w / a_{\text{lat}}}$. With $w = 1.0$\,m (half-width of the pedestrian zone), $t_{\text{maneuver}} = 0.71$\,s, giving a total cyclist swerve time of $t_{\text{swerve}} = t_{\text{react}} + t_{\text{maneuver}} = 0.84 + 0.71 = 1.55$\,s. This is speed-independent and faster than braking above 5\,km/h. We evaluate conservatively against the pedestrian PRT (1.87\,s) since warning the pedestrian is the system's primary function, but note that the cyclist's swerve capability provides an additional 0.32\,s safety margin not captured by our metrics.

The ground truth and the decision pipeline share the concept of closing proximity, which introduces a degree of circularity. We mitigate this in three ways. First, the ground truth uses trajectory-level CPA and TTC computed from full scenario knowledge, while the pipeline operates causally with noisy per-frame observations. Second, the ground truth applies a stopping-distance criterion that the pipeline does not check directly. Third, we report a sensitivity analysis over the three free thresholds in the ground-truth definition (\cref{tab:gt_sensitivity}), showing that the headline results are stable across a factor-of-two variation in each parameter.

\begin{table}[t]
  \caption{Sensitivity of results to ground-truth labeling thresholds (with fisheye localization error). Each row perturbs one parameter while holding the others at the default.}
  \label{tab:gt_sensitivity}
  \centering
  \small
  \begin{tabular*}{\columnwidth}{@{}l@{\extracolsep{\fill}}lcc@{}}
    \toprule
    Parameter & Value & Sensitivity & SevFN \\
    \midrule
    CPA radius & 3\,m   & 91.4\% &  8.9\% \\
    (default 5\,m) & 5\,m   & 93.3\% &  7.6\% \\
                   & 7\,m   & 92.5\% &  8.4\% \\
    \midrule
    Stop. margin & 60\%  & 91.9\% &  9.5\% \\
    (default 80\%) & 80\%  & 93.3\% &  7.6\% \\
                   & 100\% & 94.0\% &  6.3\% \\
    \midrule
    TTC threshold & 2.0\,s & 93.1\% &  6.7\% \\
    (default 3.0\,s) & 3.0\,s & 93.1\% &  7.7\% \\
                     & 4.0\,s & 92.1\% &  8.2\% \\
    \bottomrule
  \end{tabular*}
\end{table}

\subsection{Parameter Selection}
\label{sec:optimization}

The pipeline has five parameters: cyclist memory length $N$ (frames), proximity bounds $d_{\min}$ and $d_{\max}$ (m), minimum cyclist displacement $\Delta_{\min}$ (m over $k$ frames), and lookback depth $k$ (frames). \Cref{tab:pipeline_results} compares four configurations. The optimizer converged to $N\!=\!58$, $d\!=\![1.9, 24.8]$\,m, $\Delta_{\min}\!=\!0.147$\,m, $k\!=\!2$. The large $d_{\max}$ reflects the field-measured braking deceleration of 1.96\,m/s$^2$, which produces a stopping distance of 24.7\,m at 30\,km/h (vs.\ 31.0\,m with AASHTO design values of $t_{\text{react}}\!=\!2.5$\,s, $a\!=\!3.4$\,m/s$^2$~\cite{aashto2018geometric}). The cyclist memory of $N\!=\!58$ frames (1.9\,s) balances track persistence against stale alerts, and $d_{\min}\!=\!1.9$\,m filters out pairs that are already within arm's reach where avoidance is no longer possible.

\begin{table}[t]
  \caption{Parameter configuration comparison across 24 conformance scenarios with fisheye localization error. Each detected position is perturbed by the systematic error from projecting the 3D bounding box bottom-center through the equidistant fisheye model.}
  \label{tab:pipeline_results}
  \centering
  \small
  \resizebox{\columnwidth}{!}{%
  \begin{tabular}{@{}lcccc@{}}
    \toprule
    Configuration & Sens & Spec & SevFN & Fatigue \\
    \midrule
    Narrow window ($d_{\max}\!=\!10$\,m) & 56.1\% & 97.4\% & 61.5\% & 19.6\% \\
    Conservative                         & 77.1\% & 92.6\% & 31.0\% & 28.9\% \\
    Speed adaptive                       & 93.1\% & 91.8\% &  9.4\% & 33.7\% \\
    Selected                             & 93.3\% & 92.3\% &  7.6\% & 32.3\% \\
    \bottomrule
  \end{tabular}}
\end{table}

\paragraph{Ablation: effect of localization error.} \Cref{tab:ablation_noise} isolates the impact of fisheye localization error by comparing results with and without the systematic position offset from the bbox bottom-center projection. With the field-measured braking parameters, the localization error has a modest effect: sensitivity changes by less than 1 percentage point. The 2-frame lookback ($k\!=\!2$) smooths out position offsets effectively.

\begin{table}[t]
  \caption{Ablation: effect of fisheye localization error on the Selected configuration ($N\!=\!58$, $d\!=\![1.9, 24.8]$\,m, $k\!=\!2$).}
  \label{tab:ablation_noise}
  \centering
  \small
  \resizebox{\columnwidth}{!}{%
  \begin{tabular}{@{}lcccc@{}}
    \toprule
    Condition & Sens & Spec & SevFN & Fatigue \\
    \midrule
    Without localization error & 91.9\% & 92.5\% &  8.4\% & 31.9\% \\
    With localization error    & 93.3\% & 92.3\% &  7.6\% & 32.3\% \\
    \bottomrule
  \end{tabular}}
\end{table}

\subsection{Size-Aware Stochastic Evaluation}

\begin{figure}[t]
  \centering
  \includegraphics[width=\linewidth]{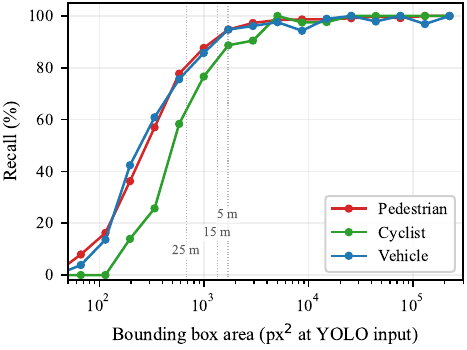}
  \caption{Recall vs.\ projected bounding box area at YOLO input resolution.}
  \label{fig:ar_curve}
\end{figure}

Under real-world conditions, detection recall depends on the projected object size in the fisheye image. For each agent, we project its 8 three-dimensional bounding box corners through the camera model and compute the tightest enclosing rectangle at the YOLO input resolution. \Cref{fig:ar_curve} shows the measured recall as a continuous function of this area, obtained from the fisheye-augmented validation set. On each Monte Carlo frame, each agent is dropped with probability $1 - \text{AR}(a)$, where $a$ is the projected area interpolated from this curve.

\subsection{Deployment Design-Space Analysis}
\label{sec:deployment_analysis}

NYC crosswalks span 30--60 feet of road width. We evaluate one-camera and two-camera deployments with cameras on traffic signal poles at opposite crosswalk edges. Two-camera fusion uses independent detection: $P_{\text{fused}} = 1 - (1 - \text{AR}_1)(1 - \text{AR}_2)$.

\begin{figure}[t]
  \centering
  \includegraphics[width=\linewidth]{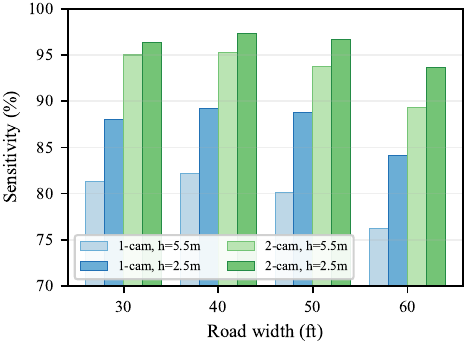}
  \caption{Monte Carlo sensitivity for one- and two-camera deployments.}
  \label{fig:deployment}
\end{figure}

\Cref{fig:deployment} shows the Monte Carlo sensitivity across road widths. Two-camera fusion raises sensitivity by 10--13 percentage points over single-camera configurations.

\subsection{Estimated Warning Budget}
\label{sec:warning_budget}

The system produces an audible and visual alert when a closing cyclist is detected. We define the \emph{estimated warning budget} as the time from the first alert to the cyclist's closest approach. This is a geometric estimate computed from the scripted trajectories; whether a real pedestrian would notice, interpret, and act on the specific light and bell signals within this time has not been tested in a controlled study and remains an open question.

To contextualize the estimated budget, we compare it against three pedestrian perception-reaction time (PRT) thresholds measured at signalized crosswalks~\cite{ped_prt_gait} (combined-sex averages):
\begin{enumerate}
  \item \textbf{Anticipating light change} (0.77\,s): pedestrian expecting the signal.
  \item \textbf{Looking straight ahead} (0.84\,s): attentive, watching the signal.
  \item \textbf{Distracted} (1.87\,s): not attending to the signal.
\end{enumerate}

\noindent These thresholds were measured for standard crosswalk walk signals and not for the custom alerts used in this system. We adopt them as reference points.

\begin{figure}[t]
  \centering
  \includegraphics[width=\linewidth]{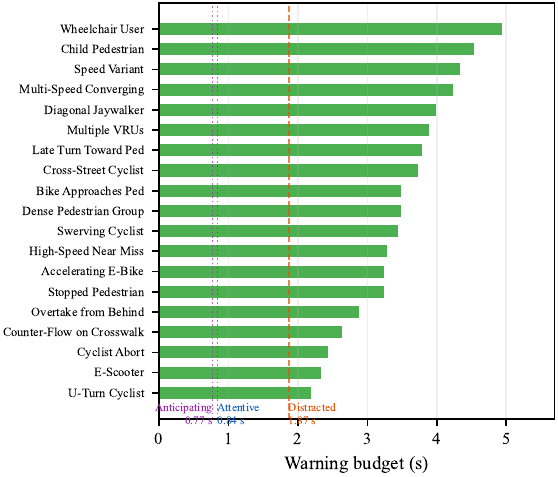}
  \caption{Estimated warning budget at alert onset for each danger scenario, compared to pedestrian PRT thresholds~\cite{ped_prt_gait}. Green: exceeds distracted PRT (1.87\,s). Orange: exceeds attentive PRT (0.84\,s). Red: below attentive PRT.}
  \label{fig:warning_budget}
\end{figure}

\Cref{fig:warning_budget} shows the estimated budget for each of the 19 scenarios with actionable danger frames at zero camera latency, with fisheye localization error applied to all positions. The mean estimated warning budget is 3.3\,s. One scenario (Swerving Cyclist) produces a zero-budget onset where the alert triggers at closest approach; the remaining 18 all exceed the attentive-pedestrian PRT (0.84\,s).

\paragraph{Failure vignette: Swerving Cyclist.} The zero-budget onset in the Swerving Cyclist scenario illustrates the value of the conformance artifact. The cyclist approaches on a straight path, then swerves sharply toward the pedestrian at close range. The pairwise closing check does not trigger until the trajectory has already curved inward, by which point the closest approach is immediate. This failure was discovered during design-time evaluation, not during field deployment. The testbench exposes it as a named, reproducible scenario that any future parameter change or rule modification must re-pass. Without the conformance suite, this failure mode would surface only as an unexplained missed alert in the field.

\begin{table}[t]
  \caption{Residual risk summary. Hazards not fully addressed by the current system.}
  \label{tab:residual_risk}
  \centering
  \small
  \resizebox{\columnwidth}{!}{%
  \begin{tabular}{@{}llll@{}}
    \toprule
    Failure mode & Scenario coverage & Sensor model & Field-blocking \\
    \midrule
    Late swerve toward ped & Yes (Swerving Cyclist) & Yes & No (detected, 0\,s budget) \\
    Occluded emergence $<$9\,m & Yes (Occluded Emergence) & Yes & No (imminent tier) \\
    Cyclist at $>$40\,km/h & Yes (Fast Approach, e-bike) & Yes & No (budget 2.3\,s) \\
    Low-light / rain & No & No & Yes \\
    Multi-cyclist occlusion & Partial (Multi-Speed) & AR curve only & Monitor \\
    Sensor tamper / failure & No & No & Yes \\
    Wheelchair on slope & No & No & Monitor \\
    \bottomrule
  \end{tabular}}
\end{table}

\Cref{tab:residual_risk} summarizes known residual risks. Hazards marked ``field-blocking'' require additional evidence before unsupervised deployment. Hazards marked ``monitor'' are partially covered but warrant targeted data collection.

\subsection{Sensor Latency Sensitivity}
\label{sec:latency}

Three distinct latency components affect end-to-end system performance:
\begin{itemize}
  \item \textbf{Processing latency} (30.1\,ms): YOLO inference, tracking, and decision logic. Fixed for a given hardware configuration.
  \item \textbf{Camera latency} (hardware-dependent): sensor readout, USB transfer, and internal buffering~\cite{crosswalk_warning_system}.
  \item \textbf{End-to-end latency}: the sum of processing and camera latency.
\end{itemize}

Processing latency is measured and fixed. Camera latency varies with the capture pipeline and is the dominant unknown in infrastructure deployments. For USB3 cameras with MJPEG capture at 30\,fps, the sensor-to-host transfer contributes one frame period (33\,ms) plus USB buffering, yielding a total camera latency of 50--100\,ms under normal conditions. V4L2 multi-buffer queuing, MJPEG decode overhead, and thermal throttling on edge devices can increase this to 150--200\,ms. We adopt 200\,ms as a realistic upper bound for a co-located camera and compute unit connected via USB3. Higher latencies arise when the camera and compute unit are separated by a network link. IP cameras streaming H.264 or H.265 over RTSP introduce encode, packetization, transport, and decode stages that can add substantial and difficult-to-characterize latency depending on the encoder GOP structure, transport protocol (UDP vs.\ TCP), and network conditions. Wi-Fi backhaul, common in temporary or portable deployments, adds further jitter. We sweep the full 0--500\,ms range to cover both co-located USB deployments and networked camera architectures.

We sweep camera latency from 0 to 500\,ms in 33\,ms steps across all 24 scenarios. To compensate, we evaluate a first-order kinematic predictor:
\begin{equation}
  \hat{\mathbf{p}}_{t} = \mathbf{p}_{t - \delta} + \delta \cdot \mathbf{v}_{t - \delta},
  \label{eq:forecast}
\end{equation}
where $\delta = n_{\text{delay}} / f_{\text{fps}}$ is the delay in seconds and $\mathbf{v}$ is the exponentially smoothed velocity in m/s. In the implementation, velocity is stored as displacement per frame and the prediction multiplies by $n_{\text{delay}}$ directly. This executes in $O(N)$ time for $N$ tracks. We also evaluate a second-order predictor that adds a $\tfrac{1}{2}\mathbf{a}\delta^2$ term.

The second-order predictor performs worse than first-order on every metric. The acceleration estimate amplifies measurement noise through the quadratic term. On non-linear scenarios such as swerving and late turns, it overshoots the true trajectory. This finding cautions against increasing prediction order without commensurate improvements to the state estimator.

\begin{figure}[t]
  \centering
  \includegraphics[width=\linewidth]{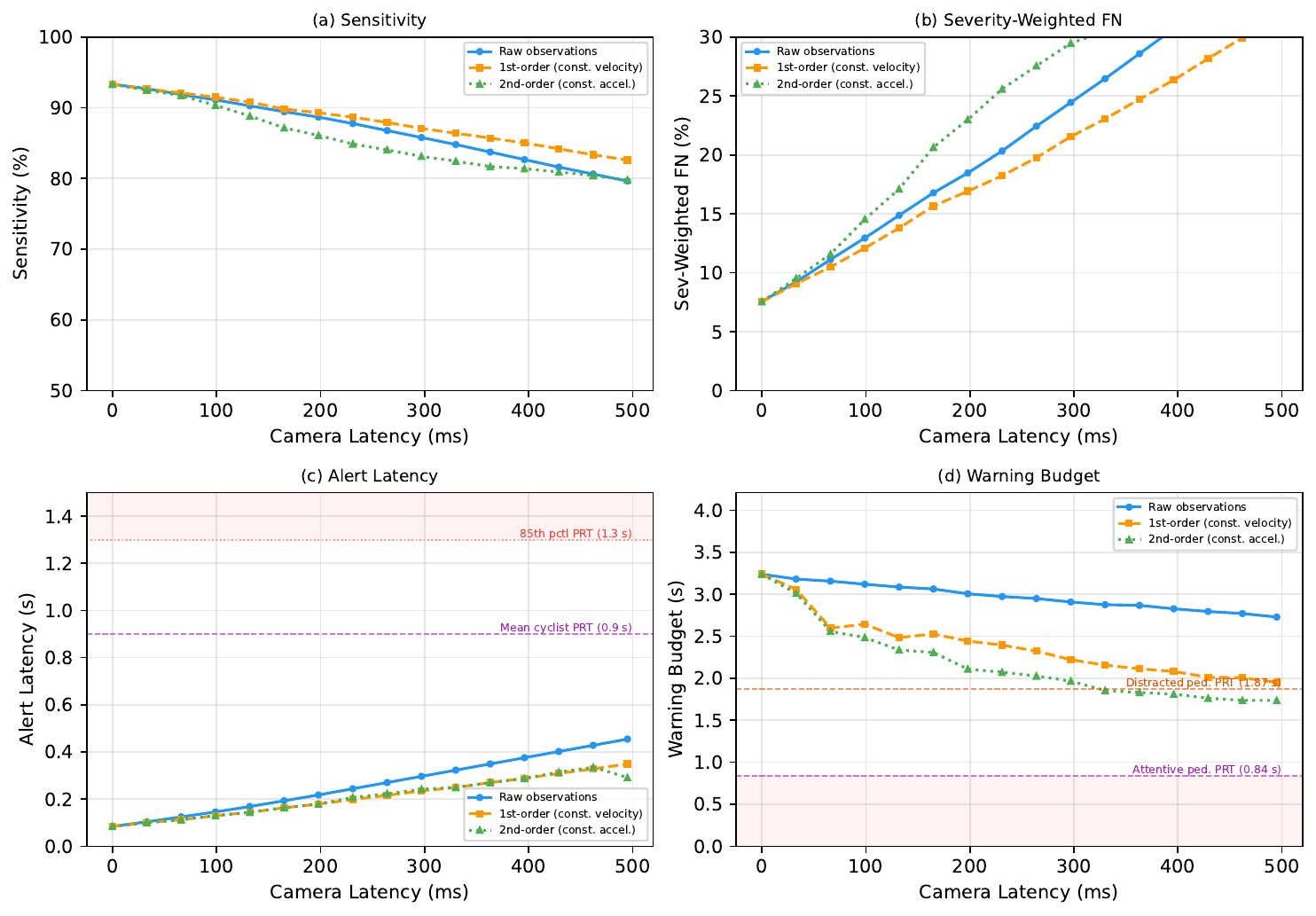}
  \caption{Conformance metrics under camera latency (0--500\,ms) for three prediction strategies. Reference lines: cyclist PRT~\cite{cyclist_prt_asce2025} and pedestrian PRT~\cite{ped_prt_gait}. Stopping distance uses 85th-percentile field values~\cite{cyclist_prt_asce2025}; AASHTO design values~\cite{aashto2018geometric} are reported separately in the braking profile comparison.}
  \label{fig:latency}
\end{figure}

At the 200\,ms co-located bound, the first-order predictor achieves 89.3\% sensitivity with a mean warning budget of 2.44\,s, providing a 0.57\,s margin above the distracted-pedestrian threshold. Under the networked 500\,ms condition, the mean warning budget still exceeds the threshold (1.95\,s vs.\ 1.87\,s), but the margin narrows to 0.08\,s, illustrating why camera latency is the dominant engineering constraint for infrastructure warning systems. \Cref{fig:latency} reports the full sweep.

\subsection{Camera Placement Optimization}
\label{sec:cam_optimization}

\begin{figure*}[t]
  \centering
  \includegraphics[width=\linewidth]{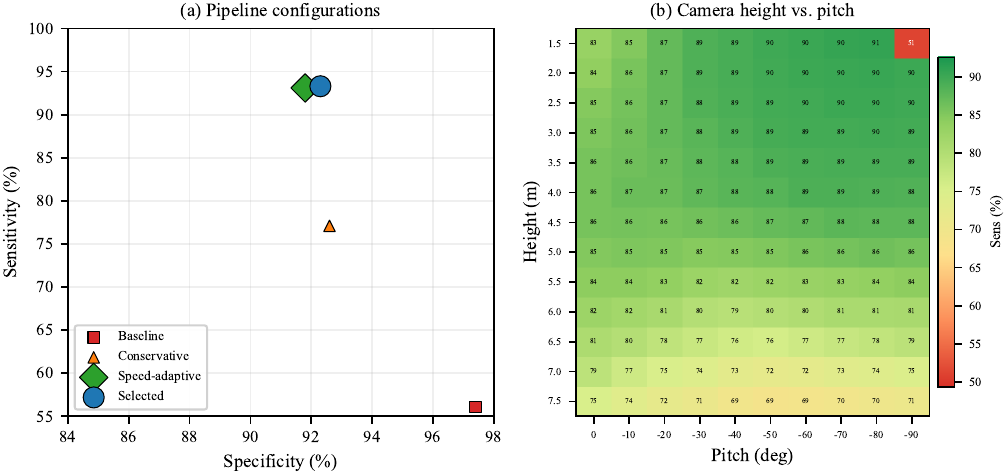}
  \caption{(a) Sensitivity vs.\ specificity for four pipeline configurations. (b) Monte Carlo sensitivity as a function of height and pitch. The dashed box marks the recommended operating region.}
  \label{fig:combined}
\end{figure*}

The mounting height and pitch affect projected object size and therefore recall. \Cref{fig:combined}(b) shows the result of a joint grid search over height from 1.5\,m to 7.5\,m in 0.5\,m steps and pitch from $0^\circ$ to $-90^\circ$ in $10^\circ$ steps on a 40-foot road (20 Monte Carlo trials per cell). Sensitivity peaks at low heights with steep downward pitch (90.6\% at 1.5\,m, $-80^\circ$) and degrades above 5\,m as objects shrink. The gradient flattens in the 2.0--3.5\,m range.

%% file: sec/5_deployment.tex
\section{Prototype Demonstration and Stakeholder Feedback}
\label{sec:deployment}

The hardware, detection model, and tracking configuration are described in \cref{sec:system}. This section reports observations from the deployed prototype and the stakeholder engagement process.

\subsection{Community-Aided Design Workshops}

The system was presented in two community-aided design workshops at Columbia University. The decision testbench was presented alongside the physical system, allowing community members and policymakers to inspect the alert logic.

Feedback from these sessions informed two design decisions: the choice of combined audible and visual alert modalities, and the decision to expose the pipeline state diagram as a stakeholder-inspectable and modifiable interface.

%% file: sec/6_conclusion.tex
\section{Conclusion}
\label{sec:conclusion}

We presented a prototype collision warning system for pedestrians and cyclists at urban intersections, running on a single edge device with a wide-angle fisheye camera. We developed a calibration pipeline that handles the corner-detection and optimizer-convergence challenges of ultra-wide lenses, and trained a fisheye-augmented detector that achieves usable recall at the frame rates required for real-time alerting. We showed that the pairwise historical closing check improves specificity over naive closing while preserving comparable sensitivity, and improves sensitivity over TTC-based rules at the cost of lower specificity.

We evaluated the system through a design-time conformance simulation with 24 hazard scenarios, providing structured evidence that the pipeline handles the enumerated encounter types including non-linear cyclist trajectories. We showed that a first-order kinematic predictor is both sufficient and preferable to a second-order predictor, and that the mean warning budget remains above the distracted-pedestrian reaction time across realistic camera latencies. We formalized the decision layer as a contestable governance artifact: stakeholders can challenge a rule by proposing a new scenario or tighter threshold and rerunning the conformance suite, and a residual risk register (\cref{tab:residual_risk}) identifies uncovered hazards that block unsupervised deployment.

Field demonstrations confirmed the expected behavior in live conditions. The relationship between warning budget and actual pedestrian response in uncontrolled settings remains an open question. The complete codebase, calibration pipeline, and scenario definitions are provided for reproducibility.

\section*{Acknowledgements}
\label{sec:acknowledgements}

This work was supported by the NSF Engineering Research Center for Smart Streetscapes under Award EEC-2133516. 